\pdfoutput=1

\documentclass[11pt]{article}

\usepackage{authblk}
\usepackage{EACL2023}

\usepackage{times}
\usepackage{latexsym}

\usepackage[T1]{fontenc}

\usepackage[utf8]{inputenc}

\usepackage{microtype}

\usepackage{inconsolata}

\usepackage{enumitem}
\setlist[itemize,1]{left=0.2em, label=$\bullet$, itemsep=6pt}

\usepackage{graphicx,subfig}
\usepackage{float}

\usepackage{xspace}

\newcommand{\vago}[0]{\texttt{VAGO}\xspace}
\newcommand{\vagon}[0]{\texttt{VAGO-N}\xspace}
\newcommand{\cats}[0]{\texttt{CATS}\xspace}
\newcommand{\oupn}[0]{\texttt{PPN}\xspace}
\newcommand{\xgboost}[0]{\texttt{XGBoost}\xspace}

\newcommand{\roberta}[0]{\texttt{RoBERTa}\xspace}

\newcommand{\bert}[0]{\texttt{BERT}\xspace}

\newcommand{\camembert}[0]{\texttt{CamemBERT}\xspace}

\newcommand{\shap}[0]{\texttt{SHAP}\xspace}

\newcommand{\tfidf}[0]{\texttt{TF-IDF}\xspace}

\usepackage{hyperref}
\title{Exposing propaganda: an analysis of stylistic cues comparing human annotations and machine classification}

\author[1,2]{Géraud Faye}
\author[3,4]{Benjamin Icard}
\author[5]{Morgane Casanova}
\author[6]{Julien Chanson}
\author[4,7]{François Maine}
\author[8]{\\ François Bancilhon}
\author[1]{Guillaume Gadek}
\author[5]{Guillaume Gravier}
\author[3]{Paul \'Egré}

\vspace{0.2in}
\affil[1]{\textit{Airbus Defence and Space}, France}

\affil[2]{\textit{Universit\'e Paris-Saclay, CentraleSup\'elec}, MICS, France}

\vspace{0.03in}
\affil[3]{\textit{Institut Jean-Nicod}, CNRS, ENS-PSL, EHESS, France}

\affil[4]{\textit{LIP6}, CNRS, Sorbonne Université, France}

\vspace{0.03in}
\affil[5]{\textit{Université de Rennes, CNRS, Inria}, IRISA, France}

\vspace{0.03in}
\affil[6]{\textit{Mondeca}, France}

\vspace{0.03in}
\affil[7]{\textit{Freedom Partners}, France}

\vspace{0.03in}
\affil[8]{\textit{Observatoire des Médias}, France}

\begin{document}
\maketitle

\begin{abstract}
This paper investigates the language of propaganda and its stylistic features. It presents the \oupn dataset, standing for Propagandist Pseudo-News, a multisource, multilingual, multimodal dataset composed of news articles extracted from websites identified as propaganda sources by expert agencies. A limited sample from this set was randomly mixed with papers from the regular French press, and their URL masked, to conduct an annotation-experiment by humans, using 11 distinct labels. The results show that human annotators were able to reliably discriminate between the two types of press across each of the labels. We propose different NLP techniques to identify the cues used by the annotators, and to compare them with machine classification. They include the analyzer \vago to measure discourse vagueness and subjectivity, a \tfidf to serve as a baseline, and four different classifiers: two \roberta-based models, \cats using syntax, and one \xgboost combining syntactic and semantic features.

\end{abstract}


\section{Introduction}

In times of warfare as well as in authoritarian regimes, state propaganda is an informational weapon whose aim is to damage the opponents' reputation and to maintain trust in the state's actions \cite{jowett2019propaganda}.
With the development of the internet and social networks, propaganda has new media to sprawl and to cross borders \cite{da-san-martino-etal-2020-semeval}.
Current trends on news consumption show an increase in the number of people getting informed on digital device.\footnote{\url{https://www.pewresearch.org/journalism/fact-sheet/news-platform-fact-sheet/}} Internet platforms are a new playground for propagandists, where they can disseminate partisan pieces among news articles and opinions shared on social media.

The rhetorical techniques of propagandists differ and their detection is currently a topic of interest~\cite{ martino_propaganda_2020,Quaranto2021propaganda}. In this paper, we pursue this general line of analysis, by examining the language of propaganda and its stylistic features. More specifically, we propose a comparison between human classification and machine classification of propaganda.

We present the \oupn dataset, standing for Propagandist Pseudo-News, a multisource, multilingual, multimodal dataset composed of news articles extracted from websites identified as propaganda sources by \textcolor{darkblue}{Newsguard} and \textcolor{darkblue}{Viginum}, a French state-backed misinformation and foreign interference surveillance organisation. Composition of the dataset is detailed in Section~\ref{sec:corpus_desc}.

To analyse the corpus and deepen our understanding of the language of propaganda, we also conducted a multilabel annotation experiment involving randomly mixing articles from that corpus with a sample of articles from mainstream French newspapers. The experiment is detailed in Section~\ref{sec:materials}, and the results are presented in Section~\ref{sec:annotations}, showing that regular press articles and articles from the corpus are recognizably different to annotators, despite sharing topics.



To find the cues characteristic of each corpus, we then used different techniques. In Section~\ref{sec:vago}, we use the expert system \vago to check on the occurrence of subjective and vagueness markers in either type of corpus, since intentional vagueness \cite{Egre&Icard2018} is among recognized techniques of propaganda \cite{martino_propaganda_2020} and its higher prevalence detectable in fake news \cite{Guelorget2021combining}. Then in Section~\ref{sec:ml}, we train machine learning models to detect articles from propagandist sources, three based on text processing and one on stylistic and syntactic features. Explainability capabilities of the models are used to confirm the features learnt by the models and to discuss ways in which they can be improved.

\section{The \oupn dataset}\label{sec:corpus_desc}

The proposed \oupn dataset is diverse in terms of sources, topics and used languages. The corpus has been extracted from 5 sources (news distribution by source is shown in Table~\ref{tab:distrib_source}), all of which were created after the Russian invasion of Ukraine on February 24, 2022:

\begin{itemize}
    \item \textcolor{darkblue}{rrn.media}: \textit{Reliable Recent News} (previously named \textit{Reliable Russian News}) has the form of a news website publishing articles containing a pro-Russia or anti-Occident stance. The website contains news in 9 languages (Arabic, Chinese, English, French, German, Italian, Russian, Spanish and Ukrainian), which receive a different coverage over time. 
    
    \item \textcolor{darkblue}{tribunalukraine.info}: this website aims at accusing Ukraine of committing war crimes and financially benefiting from the conflict. The writing style is more aggressive than \textit{rrn}, as it aims at damaging Ukraine's reputation. All articles from this source are available in English, French, German, Russian and Spanish.
    \item \textcolor{darkblue}{waronfakes.com}: the counterpart of tribunalukraine, it aims at denying Russian war crimes allegations. It does not publish news articles, but short summaries of allegations, and as such it qualifies as fake news. All \textit{``debunked"} facts are available in Arabic, Chinese, English, French, German and Spanish.
    \item \textcolor{darkblue}{notrepays.today} and \textcolor{darkblue}{lavirgule.news}: these French-writing websites publish polarizing news with the aim of damaging trust in Western institutions. Contrarily to the first three sources, which were created at the beginning of the Russian invasion, \textit{notrepays} and \textit{lavirgule} were created one year later, with a related agenda.
\end{itemize}

\begin{table}[]
    \centering
\begin{small}
    \begin{tabular}{p{2.2cm}|c}
        Source & Number of documents \\
        \hline
        rrn & 12,427 \\
        tribunalukraine & 4,975 \\
        waronfakes & 344 \\
        notrepays & 480 \\
        lavirgule & 503
    \end{tabular}
\end{small}
    \caption{PPN articles distribution by source.}
    \label{tab:distrib_source}
\end{table}

Unlike some previous publications~\cite{heppell2023analysing}, we present the propaganda articles in their original language for analysis, but knowing that several of the sites present translations in different languages.
We share the collected dataset on the following GitHub repository: \url{https://github.com/hybrinfox/ppn}. The distribution of articles by languages is shown in Table~\ref{tab:distrib_lang}. 

\begin{table}[]
    \centering
\begin{small}
    \begin{tabular}{p{2.2cm}|c}
        Language & Number of documents \\
        \hline
        Arabic & 1,079 \\
        Chinese & 794 \\
        English & 3,219 \\
        French & 4,141 \\
        German & 3,341 \\
        Italian & 1,796\\
        Russian & 1,435\\
        Spanish & 2,485\\
        Ukrainian & 439
    \end{tabular}
\end{small}

    \caption{PPN articles distribution by language.}
    \label{tab:distrib_lang}
\end{table}

\section{Annotated corpus and labels}\label{sec:materials}

To understand how propaganda can be perceived and its characteristics, we conducted an annotation experiment on a subset of the French \oupn dataset. In order to balance the dataset, we added articles from five French national newspapers of different political orientations, namely  \textcolor{darkblue}{lefigaro.fr}, \textcolor{darkblue}{lemonde.fr}, \textcolor{darkblue}{marianne.fr}, \textcolor{darkblue}{liberation.fr} and \textcolor{darkblue}{mediapart.fr}. The articles were randomly selected among those sources. They had to be published after the beginning of the Ukraine invasion (February 24, 2022) and to contain at least the mention of Russia or Ukraine. An additional filter, based on article length, was applied to limit bias linked to the length of articles. All annotated articles contained between 1,000 and 10,000 characters (shorter articles belong almost exclusively to the propaganda class and longer articles always belong to the regular class). A total of 48 articles were selected for each type of press, with a maximum of 14 and a minimum of 7 articles by source in the alternative press, and a maximum of 15 vs. a minimum of 1 by source in the regular press, and roughly similar distributions across the two types.



\begin{figure}[h]
\begin{scriptsize}
\fbox{
\parbox{7.4cm}{
\begin{itemize}


\item[$\bullet$] \textcolor{black}{\textbf{Vague:}} the information contained in the article is general with few details or specific facts.


\item[$\bullet$] \textcolor{black}{\textbf{Subjective:}} the article essentially presents opinions and the explicit or implicit subjective viewpoint of its author.


\item[$\bullet$] \textcolor{black}{\textbf{Exaggeration:}} the article presents information in an exaggerated or excessive manner.


\item[$\bullet$] \textcolor{black}{\textbf{Pejorative:}} the article primarily aims to vilify individuals or institutions.


\item[$\bullet$] \textcolor{black}{\textbf{Descriptive:}} the article essentially reports facts or events rather than opinions.  


\item[$\bullet$] \textcolor{black}{\textbf{Propaganda:}} the article gives a biased presentation of the situation and seems to serve above all the interests of a state or organization.


\item[$\bullet$] \textcolor{black}{\textbf{Satirical:}} the article is intended to make people laugh and is written in a joking tone.


\item[$\bullet$] \textcolor{black}{\textbf{Dishonest Title:}} the title reports false or artificially inflated information. 


\item[$\bullet$] \textcolor{black}{\textbf{Adequate Sources:}} the article cites its sources sufficiently and accurately.


\item[$\bullet$] \textcolor{black}{\textbf{Fake News:}} in your opinion, the article deserves to be called "fake news".

\item[$\bullet$] \textcolor{black}{\textbf{False Information:}} the article contains at least one false information.


\end{itemize}}}
\end{scriptsize}
\caption{Description of the 11 labels used for the annotation task.}
\label{fig:annot-labels}
\end{figure}

Eleven labels were used for the annotations. Figure \ref{fig:annot-labels} presents them in the order in which annotators had to mark them, with a summary of their definition. The 11 labels included 5 labels targeting manipulative content proscribed by the deontology of journalism\footnote{See the 1971 \href{https://graphism.fr/wp-content/uploads/2017/03/charter-of-munich-english.pdf}{Charter of Munich}.} and the Gricean norms of cooperative discourse (Quality in particular, \citealt{grice1975logic}), namely ``Dishonest Title'', ``Fake News'', ``False Information'', ``Exaggeration'', and ``Propaganda''. We also included 2 labels  ``Satirical'' and ``Pejorative'', targeting jocular and adversarial intention; and finally, 4 labels for features susceptible to be applicable to either type of press, with 2 labels targeting the expression of opinion or its absence, namely ``Subjective'' and ``Descriptive'', and 2 labels targeting the quality of justification, namely ``Vague'' and ``Adequate Sources''. Each label was explicitly defined and accompanied by examples in the annotation manual, except for ``Fake News'', which was deliberately left up to the annotator to judge without explicit criteria, in order to find out about its best predictors among the other labels. The label ``False Information'' was presented last, since the annotators were told they had the option to do some research and fact-checking on each topic if necessary, but in order to minimize the risk of the annotators coming across the source of the articles. The labels were binary (1 for ``applies'' and 0 for ``does not apply'') and the annotators forced to choose between them (with the option of giving a free commentary). Some of our labels, finally, overlap with the propaganda techniques listed in \citet{da-san-martino-etal-2020-semeval}, in particular our label ``Pejorative'' with their ``Name calling'' and ``Doubt'', ``Exaggeration'' with ``Exaggeration/Minimization'',  ``Satirical/Pejorative/Subjective'' with their ``Loaded language'', and ``Vague'' with their ``Obfuscation/Intentional vagueness'', except that they define vagueness mostly in terms of confusion and unclarity, whereas our definition targets generality/lack of specificity.

\begin{figure}
    \centering
    \includegraphics[width=\linewidth]{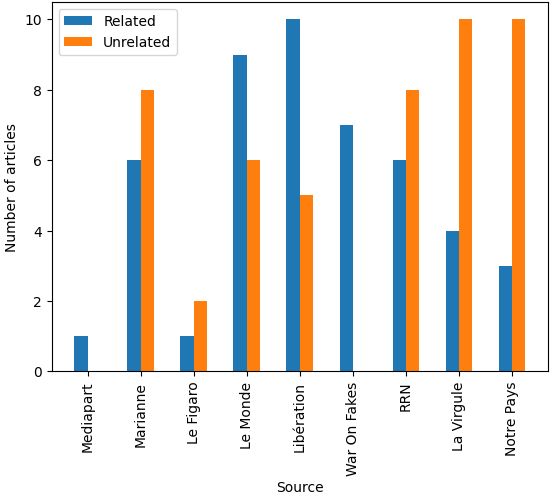}
    \caption{Topic distribution of articles from the annotated corpus.}
    \label{fig:topics_distrib}
\end{figure}

After the annotation experiment, an additional analysis of the topics was conducted to ensure that regular articles were roughly about the same topics as propaganda articles, in order to validate the experiment results. To this end, we labeled the articles depending on whether they were directly about the armed conflict (labeled \textit{Related}) or about other topics such as economic sanctions or politics (labeled \textit{Unrelated}). The articles' distribution is shown in Figure~\ref{fig:topics_distrib}.

Every source, with the exception of \textit{waronfakes} and \textit{mediapart}, had articles in both classes. \textit{mediapart} had only one article meeting our filtering conditions, and \textit{waronfakes} aims at denying war crimes allegations, so it is logical that it only contains articles directly about the armed conflict. Unexpectedly, the sample from \textit{lavirgule} and \textit{notrepays} contained more articles not directly linked to the conflict. 
Those articles seem to aim at polarizing the public debate not only on the war in Ukraine, but on other topics as well, including French politics. Overall, the annotated dataset is balanced, with 27 \textit{Related} regular articles, 21 \textit{Unrelated} regular articles, 20 \textit{Related} propaganda articles, and 28 \textit{Unrelated} propaganda articles.


\section{Analysis of the annotations}\label{sec:annotations}

The 6 annotators included the designers of the experiment. Only one of them had briefly seen the texts prior to annotating, in order to upload them on the form used for the annotation task, but without verifying their content. The articles were presented in a common random order for all participants. To avoid bias by source, the URL was removed,  in contrast to other datasets (viz. ISOT, \citealt{ahmed_detecting_2017} or \citealt{horne2017just}).



One article happened to contain mostly video links, leaving a meta-content description of the journal’s policies on cookies: it could not be annotated, and was removed, leaving a total of 48 alternative vs. 47 regular articles for analysis. Among those, five articles (4 regular, 1 alternative) happened to bear an indication of their source by self-citation in their content. Eleven articles were also truncated because they were behind a paywall (ending on the necessity to subscribe in order to access content). We kept them for analysis, but knowing that they might introduce a confound. Importantly, however, post-hoc analyses made after exclusion of those 16 articles show the same main contrasts as reported below.

The combined dataset presents individual annotations grouped by annotator, instead of aggregate data (as PolitiFact and GossipCop, \citealt{shu2018fakenewsnet}), dropping personal commentaries on the articles to secure anonymity.


In order to assess the quality of the annotations, we calculated the inter-rater agreement based on the percentage of agreement between annotators, rescaled to 0 in a case of equal split between annotators (3:3), and to 1 in case of unanimity (6:0). That is, for each document, we computed the proportion $x$ of 1-answers, rescaled by the function returning the value $|2x-1|$. For example, a value of 0.4 indicates that 70\% of the raters go in the same direction, 
while a value of .6 or above indicates 80\% of agreement or more.



\begin{figure}[t]
\subfloat[\centering Mean inter-annotator scores per label.]{\label{fig:all}\includegraphics[width=0.49\textwidth]{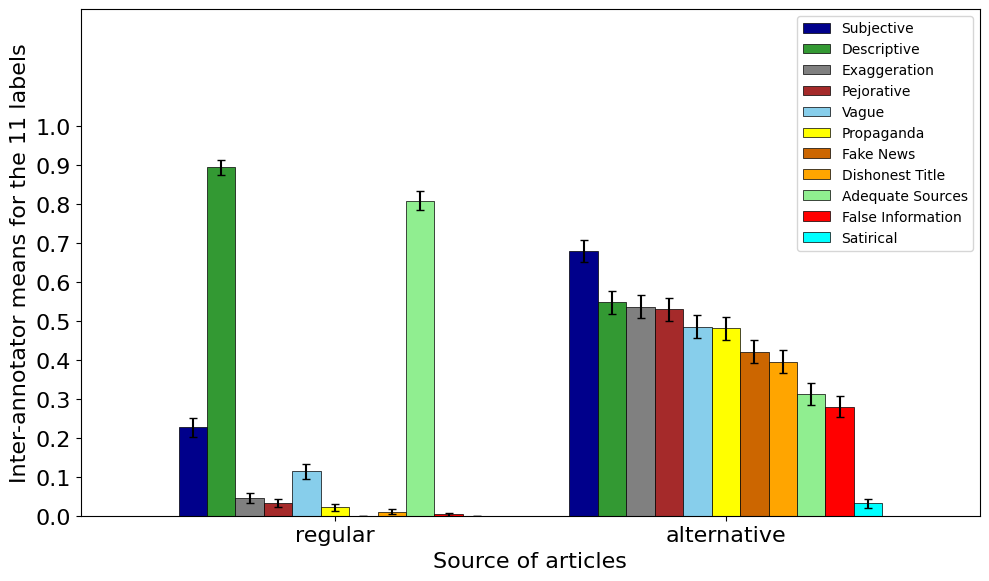}}%
\qquad
\subfloat[\centering Mean inter-annotator agreement per label.]{\label{fig:agreement}\includegraphics[width=0.49\textwidth]{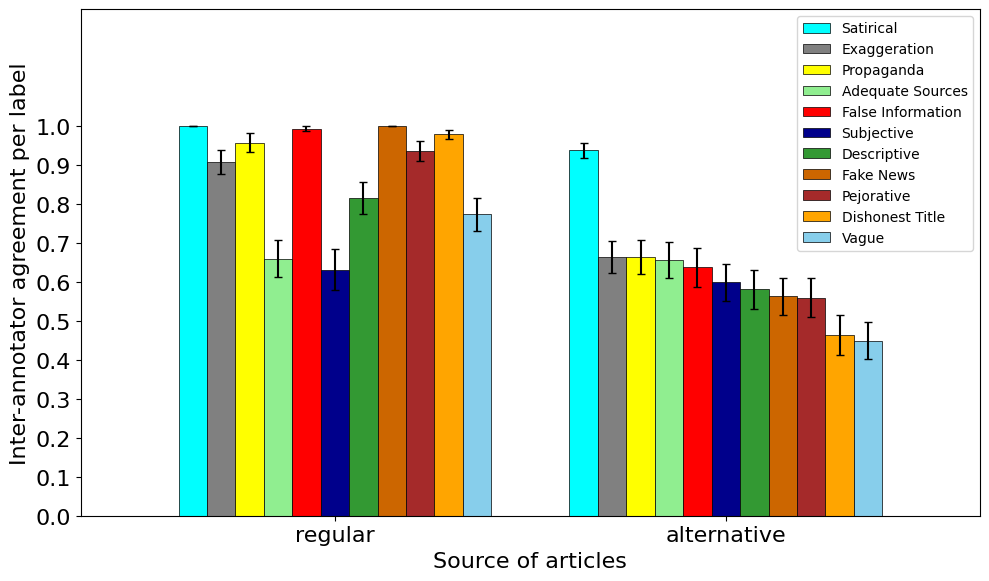}}%

\caption{\label{fig:comaprison} Mean scores and agreement by label (error bars=standard error of the mean).}
\end{figure}

As shown in Figure \ref{fig:agreement}, for both the regular and the alternative press articles, all labels reached a mean value above .4, indicative of moderate to high agreement. The agreement between annotators increases systematically from the alternative to the regular corpus, meaning that for each label, the agreement is higher in the regular corpus, compared to the alternative corpus. 

Regarding the labels themselves, Figure \ref{fig:all} shows a strong contrast between the two types of corpora. Except for the label ``Satirical'', which is almost never used in either type of corpus, the other 10 labels are used in very distinct proportions in either type of corpora (paired t-tests between the two corpora by label are all significant at the $\alpha=.01$ significance level). While each of the 10 remaining labels is applied to some extent in the alternative corpus, two labels are conspicuously never applied in the case of the regular corpus, namely: ``False Information'' and ``Fake News''. The labels ``Descriptive'' and ``Adequate Sources'', used for both types of corpora, are used in much higher proportion in the regular case. The labels ``Subjective'' and ``Vague'', while occurring for the regular corpus, are much less prevalent in the regular corpus. Finally, all other labels, in particular ``Exaggeration'', ``Propaganda'', ``Pejorative'', ``Dishonest Title'', are applied only marginally in the regular corpus.



\begin{figure}
    \centering
    \includegraphics[width=\linewidth]{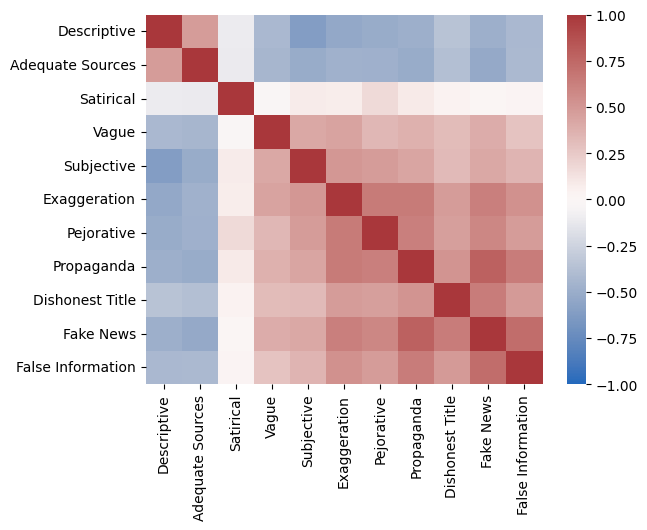}
    \caption{Correlation matrix of the 11 labels used for human annotations.}
    \label{fig:corr_mat}
\end{figure}

The correlation matrix of the labels is displayed in Figure~\ref{fig:corr_mat}.
The label ``Satirical'' is not correlated to other labels, due to its low frequency in the annotations (about 1.5\% of annotations), and is left out in the remaining of the analysis. Two main groups of labels emerge from the matrix: the labels ``Descriptive'' and ``Adequate Sources'' are strongly correlated with each other and inversely with the others, and the remaining labels, including ``Vague'', ``Subjective'', etc., are positively correlated to various degrees. Our main label of interest, ``Propaganda'', correlates most strongly with ``Fake News'', ``Pejorative'', and ``Exaggeration''.

In summary, the annotators were able to reliably discriminate between the two corpora, across each of the dimensions selected by a specific label, and moreover the strong correlation between the labels ``Propaganda'' and ``Exaggeration'' legitimizes an analysis in terms of stylistic cues.


\section{Analysis with the \vago tools}\label{sec:vago}

To see what textual features might explain the difference between the two classes, we used the lexical database and analyzer \vago \cite{icard2022vago}.
For a given text, \vago calculates three scores: a score of vagueness, a score of opinion, and a score of relative detail (compared to vagueness). To calculate the vagueness score of a text, the system checks for the occurrence of vague expressions, subcategorized into four types: generality $V_G$ (``\textit{some}'', ``\textit{or}''), approximation $V_A$ (``\textit{about}'', ``\textit{almost}''), one-dimensional vagueness $V_D$ (``\textit{old}'', ``\textit{many}''), and multi-dimensional vagueness $V_C$ (``\textit{good}'', ``\textit{effective}''). For opinion, \vago checks for the occurrence of implicit markers of subjectivity (all expressions of type $V_D$ and $V_C$, including evaluative adjectives and pejorative terms), as well as explicit markers (first-person pronouns, exclamation marks). For detail, finally, the system compares the ratio of named entities to vague terms.



While \vago does not incorporate any world-knowledge, previous studies on larger corpora have shown that the \vago scores of vagueness and opinion were positively correlated with the label ``biased'' in news articles \cite{Guelorget2021combining,icard&alWI-IAT2023}, and that the score of detail-vs-vagueness was negatively correlated with the label ``Satirical'' \cite{icard&alWI-IAT2023}. Hence, we asked if the \vago scores of vagueness, opinion, and detail might be good predictors of the human annotations, and in particular of labels such as ``Exaggeration'', ``Pejorative'', ``Propaganda'' and ``Dishonest Title''.

\begin{figure}[h]
\includegraphics[width=0.225\textwidth]{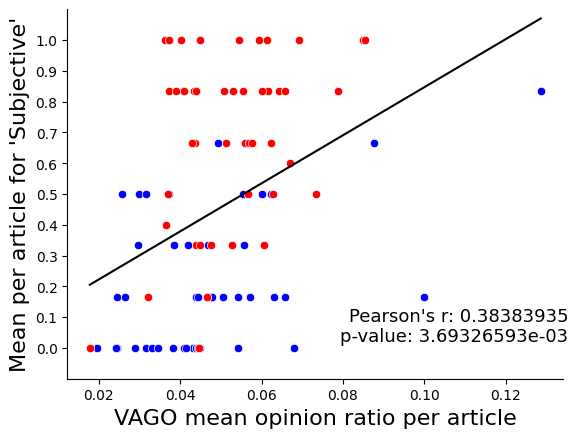}%
    \quad
\includegraphics[width=0.225\textwidth]{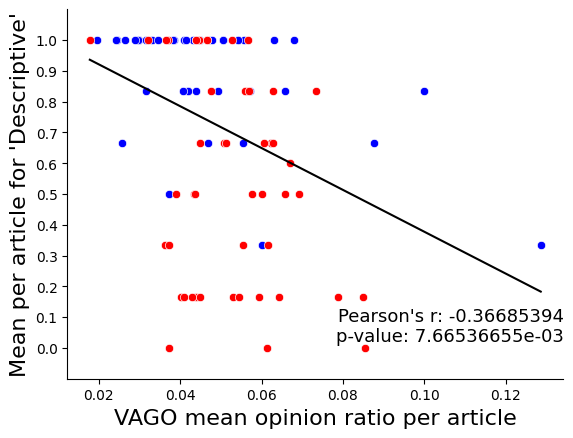}%
\caption{Pearson correlations between the \texttt{VAGO} mean opinion score per article 
and the mean scores for labels ``Subjective'' (left) and ``Descriptive'' (right). Blue data points correspond to regular articles while red data points correspond to alternative press articles.}\label{fig:plot-correls}
\end{figure}

To investigate this question, 
we calculated the correlation between the \vago scores for each article of the corpus and the mean inter-annotator scores for all of the 10 labels (``Satirical'' left aside). As shown in Table \ref{tab:correlations}, the labels ``Subjective'', ``Exaggeration'' and ``Pejorative'' turned out to be positively correlated to the \vago scores of vagueness and opinion, and negatively correlated to the scores of detail-vs-vagueness. Consistent with these results, the scores of vagueness and opinion were also negatively correlated with labels ``Descriptive'' and ``Adequate Sources''. By contrast, labels ``Propaganda'', ``Dishonest Title'', ``Fake News'' and ``False Information'' turned out to be positively correlated to the scores of vagueness only. All these correlations are weak to moderate, but they replicate results found in previous studies, with an even higher order of magnitude in the labels ``Subjective'' and ``Descriptive'' connected to \vago's opinion score, as presented in Figure \ref{fig:plot-correls}. 

Human annotations of the label ``Vague'' did not correlate with \vago scores of either vagueness or detail, however, contrary to expectations. We conjecture that this could be due to a discrepancy between the definition given of the label, which targets generality vagueness, and the fact that the \vago vagueness score is based on more types of vagueness, in particular the semantic vagueness of one-dimensional and multi-dimensional adjectives, which represent 96\% of the \vago lexicon.

Despite that, what Table \ref{tab:correlations} shows is that the \vago scores track the clustering of labels found in Figure \ref{fig:corr_mat}: the polarity of the correlations for the labels ``Descriptive'' and ``Adequate sources'' is inverse to that of the other labels. In summary, \vago scores are correlated with the separating features of the alternative vs. regular press, but they explain only part of the variance in the annotations. In the next section, we examine classification models properly in order to get further insights.

\begin{table}[h]
{\small

\begin{tabular}[t]{l|l|l|l}
\multicolumn{1}{c}{Label} & \multicolumn{1}{|c}{vague} & \multicolumn{1}{|c}{opinion} & \multicolumn{1}{|c}{detail} \\
\hline

Vague & \phantom{..} $0.163$ & \phantom{..} $0.188$ & $-0.180$ \\
Subjective & \phantom{..} $0.344^{*}$ & \phantom{..} $0.384^{**}$ & $-0.238$ \\
Exaggeration & \phantom{..} $0.282$ & \phantom{..} $0.222$ & $-0.225$ \\
Pejorative & \phantom{..} $0.289$ & \phantom{..} $0.222$ & $-0.265$ \\
Descriptive & $-0.371^{**}$ & $-0.367^{**}$ & \phantom{..} $0.228$ \\
Propaganda & \phantom{..} $0.249$ & \phantom{..} $0.165$ & $-0.152$ \\
Dishonest Title & \phantom{..} $0.257$ & \phantom{..} $0.164$ & $-0.206$ \\
Adequate Sources & $-0.210$ & $-0.210$ &
\phantom{..} $0.130$ \\
Fake News & \phantom{..} $0.233$ & \phantom{..} $0.178$ & $-0.148$ \\
False Information & \phantom{..} $0.214$ & \phantom{..} $0.140$ & $-0.099$ \\

\end{tabular}

}
\caption{Pearson correlations between the labels' mean scores and the \vago scores ($^{*}$ and $^{**}$ indicate $p$-value $<.05$ and $<.01$), with Bonferroni correction.}\label{tab:correlations}
\end{table}

\section{Machine learning for propaganda detection}\label{sec:ml}

Propaganda detection \cite{martino_propaganda_2020} from texts can be a difficult task depending on the form of the content. Classifying sentences \cite{mapes-etal-2019-divisive} is harder, even for large language models (LLMs) such as \bert \cite{devlin_bert:_2018}. In this section, a methodology for training a propaganda detection model is explained and evaluated. Smaller models with explainability capabilities were also trained in order to identify which parts of the articles the model considers when taking its decision.

\subsection{Dataset for detecting propaganda related to the conflict}

In order to train a model that could be used to identify propaganda articles, it is required to also collect regular press articles on a related topic. Here, we present the larger corpus of regular press from which the French subset of the previous section was drawn. This larger corpus also contains English articles, since the classification model is supposed to handle classification in French and in English. 




English regular articles were collected from 11 reliable news outlets, with constraints of date (being post Ukraine invasion), length (between 1,000 and 10,000 characters), and topic (mention Russia and Ukraine). English regular articles were collected using \texttt{news-please}~\cite{Hamborg2017} before being filtered. The articles distribution by source is given in Table~\ref{tab:distrib_en_regular}. The wider set of French regular articles was collected in the same way, but with a more limited choice of sources, their distribution is given in Table~\ref{tab:distrib_fr_regular}.

\begin{table}[h]
    \centering
\begin{small}

    \begin{tabular}{l|c}
    \multicolumn{1}{c}{Source} & \multicolumn{1}{|c}{Number of articles}\\
        \hline
        apnews.com & 520 \\
        cbsnews.com & 63 \\
        dailymail.co.uk & 43 \\
        cnn.com & 10 \\
        usatoday.com & 10 \\
        forbes.com & 42 \\
        foxnews.com & 5 \\
        nbcnews.com & 10 \\
        nytimes.com & 4 \\
        theguardian.com & 185 \\
        washingtonpost.com & 12 \\
        \hline
        Total & 1,004
    \end{tabular}
\end{small}
    \caption{English language regular articles distribution by source.}
    \label{tab:distrib_en_regular}
\end{table}

\begin{table}[h]
    \centering
\begin{small}

    \begin{tabular}{l|c}
\multicolumn{1}{c}{Source} & \multicolumn{1}{|c}{Number of articles}\\
        \hline
        lefigaro.fr & 3 \\
        lemonde.fr & 449 \\
        liberation.fr & 386 \\
        marianne.net & 523 \\
        mediapart.fr & 6 \\
        \hline
        Total & 1,367
    \end{tabular}
\end{small}
    \caption{French regular articles distribution by source.}
    \label{tab:distrib_fr_regular}
\end{table}


\subsection{Models}
\label{ssec:models}

Five models were chosen for propaganda detection, two in English and three in French. The English\footnote{\url{https://huggingface.co/hybrinfox/ukraine-operation_propaganda-detection-EN}} and French\footnote{\url{https://huggingface.co/hybrinfox/ukraine-operation_propaganda-detection-FR}} models are available on Huggingface-hub and can be freely downloaded and tested.

The first English model used for classification is a \roberta-base model \cite{liu2019roberta} with a classification layer using the last hidden state. For practicality, we load pre-trained English \roberta weights and fine-tune the model using the HuggingFace transformers library.


The first French model combines the ``\camembert-base'' version \cite{martin2019camembert} based on the \roberta architecture \cite{liu2019roberta} (\textit{Batch Size}=10, \textit{Learning Rate}=1e-05, \textit{Epochs}=5) with one classification layer and a BCE loss function to detect whether the articles of our French larger dataset counts as propaganda or not. 

The second French model is an \xgboost \cite{XGBoost} (Extreme Gradient Boosting) model. It is a scalable, distributed gradient-boosted decision tree. Contrarily to the other three models which process texts directly, \xgboost only takes numerical values as input. In our case it takes the following parameters: the length of the sentence, the three \vago scores (vagueness, opinion, detail), the sentiment of the sentence, positive or negative (using the HuggingFace sentiment classification model ``Monsia/camembert-fr-covid-tweet-sentiment-classification''), the number of verbs, adjectives, adverbs and nouns present in the sentence and the number of occurrences of dependencies between the words (using the spaCy python library for Natural Language Processing).\footnote{\url{https://universaldependencies.org/u/dep/all.html\#al-u-dep/nmod}} The sentence features are then aggregated by an operator. Several aggregation operators were tested and gave similar results so the sum operator was chosen.

Models applicable to both languages were tested. The first is the neurosymbolic model \cats~\cite{FAYE2023102230}. It does not use \textit{a priori} knowledge on the language except for the English syntax. It is lighter than \roberta, and has explainability capabilities that will be useful to identify what the model considers a marker of propaganda. It can also be used for other languages and results for a French version have also been reported. The second one is \tfidf, with which the texts are vectorized after removing stopwords and lemmatizing the remaining words. This representation is then processed by a random forest, predicting the class of the article.


The datasets for each language were initially split between training, validation and test using a 80/10/10 ratio with no overlap. The models were chosen on the best validation score and the reported results are on the test set, which was never used during the training procedure.

\subsection{Results}

\begin{table}[h]
\centering
\begin{small}

    \begin{tabular}{l|c|c}
    Language   & Models  & Test accuracy \\
        \hline
   English   &  \roberta & 0.997 \\
      &  \cats\texttt{- EN}& 0.953 \\
      & \tfidf\texttt{- EN} & 0.985\\
        \hline
   French   &  \camembert & 0.997 \\
    & \cats\texttt{- FR} & 0.946 \\
       & \xgboost & 0.921 \\
      & \tfidf\texttt{- FR} & 0.963\\
    \end{tabular}
\end{small}
    \caption{Test accuracies for Ukraine invasion propaganda detection models.}
    \label{tab:results}
\end{table}


The models' performances on their test sets are reported in Table~\ref{tab:results}. Propaganda detection on this specific topic is easily achieved by LLMs, and even by shallow models like \cats or \xgboost. The performance of \cats is slightly lower than \roberta's, but this is expected since it contains only 0.6 million parameters, about 200 times fewer parameters than \roberta-base with its 125 million parameters. \xgboost's performance is even lower, but the model processes high-level features of the texts, lacking other features that other models can use.



\subsection{Identified markers of propaganda}

The interest of training a smaller model like \cats on the texts is to identify which markers are learnt by this machine learning model. To this end, each token's contribution to the final decision is aggregated by sentence, enabling us to recover the most salient sentences from propaganda articles. These sentences contain more markers of propaganda and can help us understand what the model is tracking when classifying articles between propaganda and regular. 

A representative example is given in Figure~\ref{fig:expl_1}. In this example, the first underlined sentence is a case of laudatory exaggeration; the second one is pejorative, and the third is again pejorative, with even a racist insinuation. Other sentences in the text contain propagandist cues, however, making the selection hard to directly interpret. For comparison, we run \vago on the text. In this case, the scores of vagueness, opinion, and detail were 0.13, 0.08 and 0.42, respectively. The underlined items correspond to vague and subjective markers found by \vago. \vago detects several adjectives used pejoratively (``Old [Joe]'', ``trivial'' and ``simple'' in particular). It misses out on others (``smug'', ``round lost''), and on more complex syntactic markers (``even'' in  ``even a child'', ``by the way'' to introduce a derogatory and covertly racist remark). But it identifies several subjective adjectives reflecting the implicit viewpoint of the writer.


\begin{figure}[!htb]
\begin{scriptsize}
\fbox{
\parbox{7.4cm}{
\noindent

\texttt{``Round Lost Joe Biden made a rant in Warsaw \underline{about} the " unity of the West " and the" power of democracy ". \textcolor{red}{But in his own country , Vladimir Putin was more \underline{believable} .} The American president' s speech in Poland was not intended as a direct response to the Russian leader , who addressed the Federal Assembly the day before - and the entire world as well. Biden ' s national security adviser Jake Sullivan claimed it was " not a rhetorical contest with anybody " . \textcolor{red}{But the 80 - year - {old} politician ' s smug stand - up proved otherwise: he tried to confront his opponent from Moscow - and appeared to yield to him.} \underline{Old} Joe was satisfied with a 20 - minute monologue on the lawn of the Royal Castle  - by comparison , Vladimir Putin spoke for 1 hour 45 minutes. Biden' s entire message was made up of \underline{high}-pitched quotations - especially for the applause he prepared : " Democracies have become \underline{stronger} , not \underline{weaker}. \textcolor{red}{Autocracies have grown \underline{weaker} , not \underline{stronger}." Quite \underline{trivial} and as \underline{simple} as possible - so that even a child would get the point . By the way , there were a \underline{lot} of children at the President' s speech , and of \underline{different} races too.} And \underline{all} of them had Ukrainian flags - in the \underline{best} traditions of American propaganda.''}}}

\end{scriptsize}

\caption{Example of an article classified as propaganda by \cats. The sentences contributing the most to the propaganda class according to \cats are highlighted in red while the \vago vocabulary is underlined. \label{fig:expl_1}}
\label{fig:textinput}
\end{figure}

\subsection{Explainability of the \xgboost model}


\begin{figure}
    \centering
    \includegraphics[width=\linewidth]{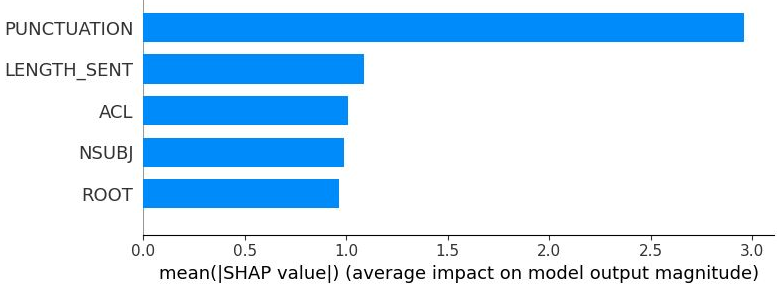}
    \caption{\shap explainability of the \xgboost model for
propaganda classification. Only the top 5 syntactic features are displayed.}
    \label{fig:shap}
\end{figure}

\begin{figure}[h]
    \centering
    \includegraphics[width=0.47\textwidth]{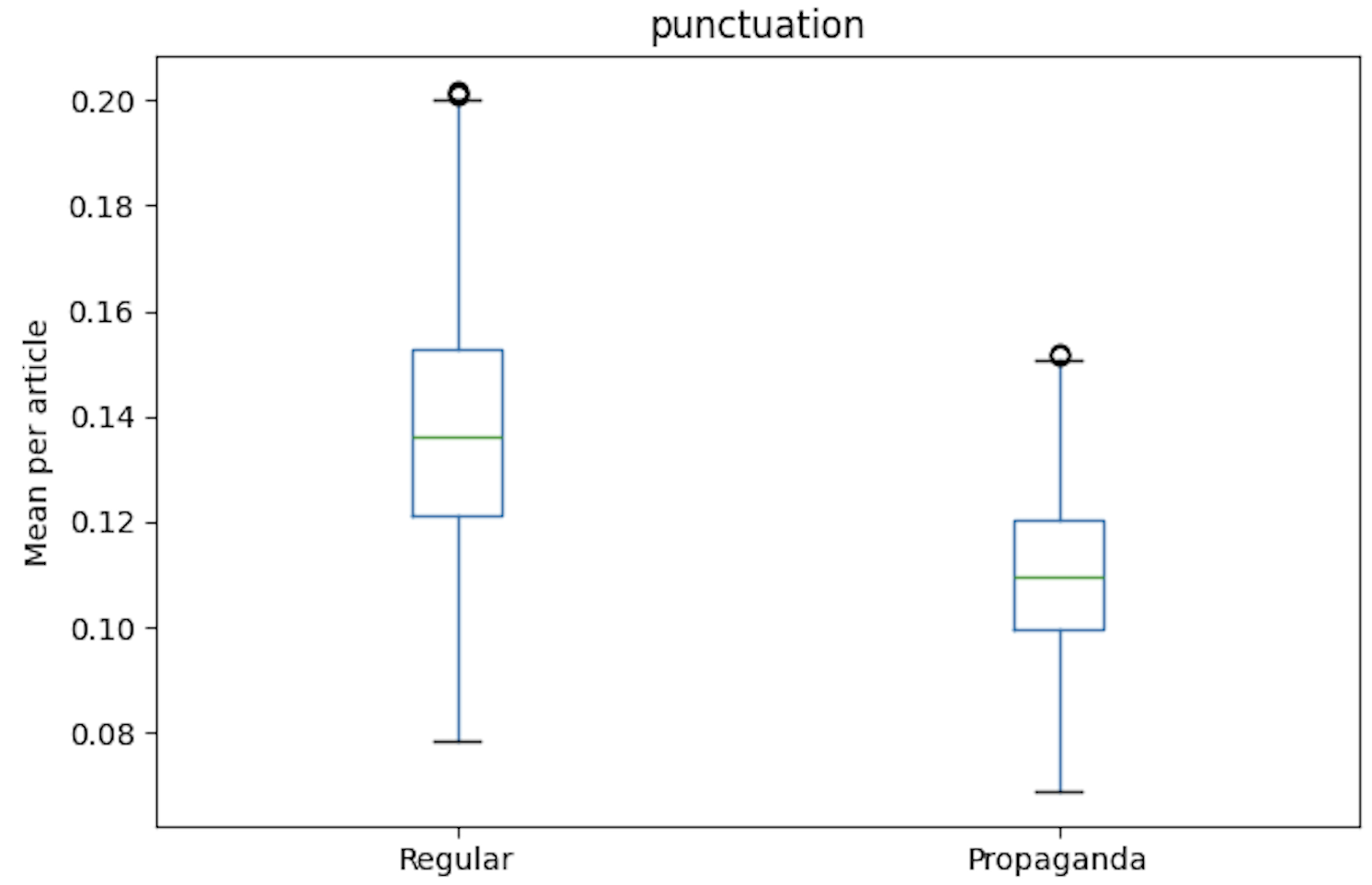}
    \caption{Percentage frequency distributions of ``punct'' dependence in regular articles vs. propaganda articles.}
    \label{fig:punct}
\end{figure}

\begin{figure}
    \centering
    \includegraphics[width=0.47\textwidth]{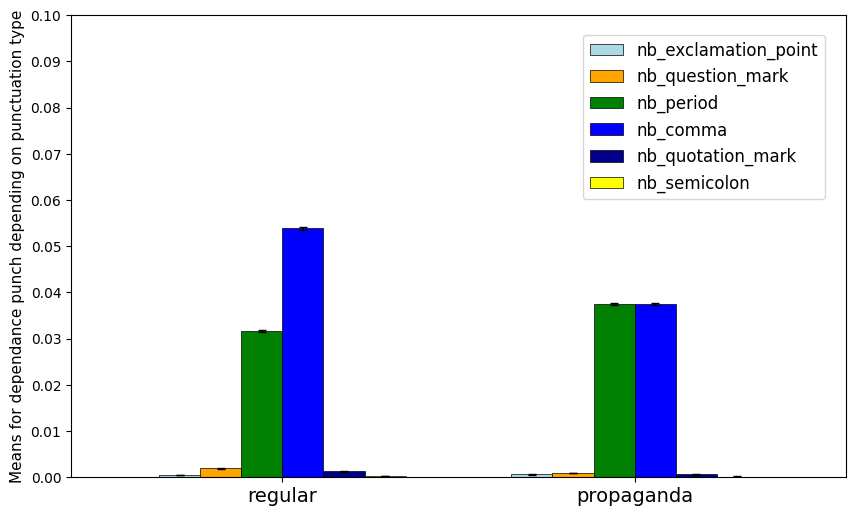}

    \caption{Relative weight of punctuation marks in either article type.}
    \label{fig:punct_plot}
\end{figure}


We used the \shap tool \cite{NIPS2017_7062} to analyze which features were the most useful for the \xgboost classification. The results are reported in Figure~\ref{fig:shap}. We observe that overall syntactic features bear more weight than other features in the detection of propaganda, with the number of punctuation marks (\texttt{PUNCTUATION}) having greater impact than the length of sentences (\texttt{LENGTH\_SENT}), the number of clausal modifiers (\texttt{ACL}), of nominal subjects (\texttt{NSUBJ}) and of sentences (\texttt{ROOT}) all receiving similar weight. 

In Figure~\ref{fig:punct}, we observe that the frequency percentage of punctuation compared to other tokens is significantly higher in regular articles than in propaganda articles ($p=8.31\times 10^{-240}$). 
We observed more precisely which type of punctuation was more represented in regular versus propaganda articles. 
Compared to other tokens, we observed that propaganda articles contain significantly more periods ($p= 2.09\times 10^{-78}$), but fewer question marks ($p= 2.12\times 10^{-32}$), fewer commas ($p= 1.47\times10^{-290}$) and fewer quotation marks ($p=1.12\times10^{-06}$) than regular articles (see Figure \ref{fig:punct_plot}). Since propaganda articles happened to be significantly shorter than regular articles ($p=7.12\times 10^{-26}$), the data was normalized by the length of the article, corresponding to the total number of tokens in the article. 

Looking at the \vagon scores on the corpora, we observe that, besides punctuation, the \vagon mean score of detail vs vagueness per article is  significantly higher for regular articles than for propaganda articles ($p= 2.66\times 10^{-44}$, with Bonferroni correction). By contrast, the differences between the \vagon scores of vagueness and opinion are no longer significant after Bonferroni correction. 

\subsection{Potential biases of machine learning models}

The near perfect accuracy of the models reported in Table \ref{tab:results} concerning Large Language Models raises questions about the shallowness of the learnt features and about potential biases in the dataset.

Regarding the first aspect, the high performance of models such as \tfidf and \cats shows that these simpler models can also detect propaganda when trained on a large dataset. The deeper models, as a result of their higher complexity, can achieve better scores, very close to 100\%.

The high accuracy of \tfidf, which uses only lexical features, manifests a clear distinction between the language of regular articles versus propaganda articles when they deal with the topic of Ukraine operation. While the models are performing well on this specific topic, there is no guarantee that they would perform equally well on other propaganda topics.

We analyzed the terms whose \tfidf scores differ significantly between the two classes in the French corpus. Among the terms more prevalent in the propaganda corpus compared to the regular corpus, we find  terms like
``état'' (\textit{state}), ``pays'' (\textit{country}), ``unis'' (\textit{united}), ``déclaré'' (\textit{declared}), ``ue'' (\textit{EU}), ``zelensky'', ``biden'', ``kiev'',``allemagne'' (\textit{Germany}), ``armes'' (\textit{weapons}). By contrast, terms like ``lire'' (\textit{read}), ``russe'' (\textit{Russian}), ``poutine'', ``kyiv'', ``invasion'', ``vladimir'', ``guerre'' (\textit{war}), ``jeudi'' (\textit{thursday}), ``mars'' (\textit{march}) and ``lundi'' (\textit{monday}) are more prevalent in the regular corpus. We notice that ``Kiev/Kyiv'' is not spelled the same way depending on the corpus. The name ``Zelensky'' is cited more in propaganda articles, whereas ``Putin'' is cited more in the regular articles of the corpus. Finally, the regular corpus contains more markers of precise time indications than the propaganda corpus, consistently with the higher \vago score of detail.

\section{Conclusion and perspectives}


In this paper, we introduced \texttt{PPN}, a multilingual propaganda dataset, and we conducted an experiment to investigate the basis on which human annotators, and then classification algorithms, can discriminate propagandist articles from non-propagandist articles on a specific topic. The annotations reveal that exaggeration, combined with lesser descriptive content, and absence of adequate sources, are prevalent in assessments of propagandist press. The \vago analyzer confirmed that the use of vague markers is significantly correlated with those features. Further analyses based on different families of classifiers revealed further syntactic cues, pertaining in particular to punctuation, but also to the lexicon. 

\newpage
Further work is needed to refine this analysis. Machine learning models, while efficient at detecting topic-specific propaganda, still have room for improvement regarding explainability and generalization to other topics. If some alignment has been observed with what humans attend to when judging an article, there is still no guarantee that language models process the text as humans would. The use of propaganda technique classifiers to identify manipulative articles yields more explainability, but at the cost of performance, especially for topic-specific propaganda.

In addition to that, while the given scores are very high, they were obtained for the task of \textit{topic-specific} propaganda detection, which is an easier task than general propaganda detection. However, 
topic-specific models still have use and can prevent the spread of disinformation in cases of conflict similar to the one used here.



While only a model for English and French propaganda detection on the Ukraine invasion is provided here, we encourage the community to use the parts of the dataset corresponding to their native language to train more classifiers. Collaborations could be considered to train a multilingual model, based on the dataset and collected regular articles from the other languages of the dataset. The same goes for annotation experiments on the way propaganda is perceived by readers, as propaganda strategies may change by languages and by target audience.



Last, in this paper we see that symbolic AI tools explain part of the classifications operated by humans as well as by classifiers. We see two ways in which explainability can be further improved: firstly, by continuing to enrich tools like \vago with lexical and even syntactic units highlighted by classifiers or by annotators in this task; secondly by considering more labels in order to improve the quality of annotations and identify more stylistic features. We introduced a label for ``Pejorative'' speech, we may also have introduced a dual label ``Laudatory'', to identify cases of glorification also typical of state propaganda, and to refine the category of ``Exaggeration''. Similarly, we may want to better control the positive and negative connotations of the labels, for instance by using labels such as ``Precise'' rather than ``Vague'', or ``Objective'' instead of ``Subjective''. 




\clearpage
\section*{Limitations}



Annotation experiments were only run on a subset of the French data. While an additional manual verification of the data quality has been done for English articles, other languages have not been manually reviewed. There may be parsing errors for some languages, and further analysis from native speakers of other languages may be required before using these parts of the dataset.

Experiments on propaganda detection were only run on two examples of Romance and Germanic languages. While language models for these types of languages are common, there is no guarantee that performant language models exist for all proposed languages from the dataset.

\section*{Ethics statement}

This article deals with the topic of propaganda and proposes a dataset to help improve propaganda detection. Proposing and sharing propaganda detection methods is crucial to keep the information space clean and safe to use for everyone.

Human exposition to propaganda should be contained. To this end, we ensured that all annotators were performing the annotation task voluntarily, with a content warning, and the possibility to stop the experiment at any time.

We encourage future works on the dataset to be conducted cautiously and on limited parts of the global dataset.

\section*{Acknowledgements}

We thank two anonymous reviewers for helpful comments and feedback. This work was supported by the programs HYBRINFOX (ANR-21-ASIA-0003), FRONTCOG (ANR-17-EURE-0017), and {PLEXUS (Marie Sk\l odowska-Curie Action, Horizon Europe Research and Innovation Programme, grant n°101086295). PE thanks Monash University for hosting during the writing of this paper.

\section*{Declaration of contribution}

All the authors contributed to the design, annotations, analysis and discussion of the results. GF, BI, MC, and PE wrote the paper, which all authors read and revised together. First authorship is equally shared between GF, BI and MC. Correspondence: geraud.faye@centralesupelec.fr, benjamin.icard@ens.fr, morgane.casanova@irisa.fr, paul.egre@ens.psl.eu.

\bibliographystyle{acl_natbib}
\bibliography{custom.bib}

\end{document}